\documentclass[journal]{IEEEtran}
\usepackage{lettrine}
\usepackage{cite}
\usepackage{amsmath,amssymb,amsfonts}
\usepackage{algorithmic}
\usepackage{textcomp}
\usepackage{xcolor}
\usepackage{graphicx}
\usepackage[linesnumbered,ruled,vlined]{algorithm2e}
\usepackage{enumitem}
\usepackage{stfloats}

\ifCLASSINFOpdf

\else

\fi

\begin{document}

\title{Efficient Sparse Artificial Neural Networks}

\author{\IEEEauthorblockN{Seyed Majid Naji\IEEEauthorrefmark{1}, Azra Abtahi\IEEEauthorrefmark{1},
		 and Farokh Marvasti\IEEEauthorrefmark{1}}
	 
\IEEEauthorblockA{\IEEEauthorrefmark{1} Department of Electrical Engineering, Sharif University of Technology, Tehran, Iran.}
\thanks{
Corresponding author: Azra Abtahi (email: azra\_abtahi@ee.sharif.edu).}}

\markboth{}%
{Shell \MakeLowercase{\textit{et al.}}: Bare Demo of IEEEtran.cls for IEEE Transactions on Magnetics Journals}

\IEEEtitleabstractindextext{%
\begin{abstract}
		The brain, as the source of inspiration for Artificial Neural Networks (ANN), is based on a sparse
		structure. This sparse structure helps the brain to consume less energy, learn easier and
		generalize patterns better than any other ANN. In this paper, two evolutionary methods for
		adopting sparsity to ANNs are proposed. In the proposed methods, the sparse structure of a
		network as well as the values of its parameters are trained and updated during the learning
		process. The simulation results show that these two methods have better accuracy and faster
		convergence while they need fewer training samples compared to their sparse and non-sparse
		counterparts. Furthermore, the proposed methods significantly improve the generalization
		power and reduce the number of parameters. For example, the sparsification
		of the ResNet47 network by exploiting our proposed methods for the image classification of ImageNet dataset uses 40\% fewer
		parameters while the top-1 accuracy of the model improves by 12\%  and 5\% compared to the dense network and their sparse counterpart, respectively. As another example, the proposed methods for the CIFAR10 dataset converge to their final structure 7 times faster than its sparse counterpart, while the final accuracy increases by 6\%. 

\end{abstract}

}

\maketitle

\IEEEdisplaynontitleabstractindextext
\IEEEpeerreviewmaketitle

\section{Introduction}

\IEEEPARstart{A}{rtificial} Neural Networks (ANN) are inspired by Natural Neural Networks (NNN) which constitute the brain. Hence, scientists are  trying to adopt natural network functionalities to the ANNs. The observations show that the  NNNs have sparse properties. First, neurons in NNNs fire sparsely in both time and spatial domains \cite{niell2008highly,harris2013cortical,mao2017sparse,mccafferty2018cortical,radosevic2019decoupling,jadhav2009sparse}. Furthermore, the NNNs profit from a   scale-free \cite{barabasi1999emergence}, small-world \cite{watts1998collective}, and sparse structure  which means each neuron connects to a very limited set of other neurons. In other words, the connections in a natural network is a very small portion of the possible connections \cite{hofer2011differential,jirsa2007handbook, clune2013evolutionary}. This property eventuates in very important features of NNNs such as low complexity, low power consumption, and  stronge generalization power \cite{safaryan2017nonspecific,ganguli2012compressed}. On the other hand, conventional ANNs do not have sparse structures which lead to the important differences between them and their natural counterparts. They  have an extraordinary large number of parameters and need  extraordinary large storage space, large sample number, and powerful underlying hardware \cite{bullmore2009complex}. 
Although deep learning has gained enormous success in wide variety of domains such as  image and video processing\cite{miao2019improved,litjens2017survey,moen2019deep}, Biomedical processing\cite{nelson2019integrating,wang2019predicting,miao2019improved,chen2019deep,amin2019evaluation,eraslan2019deep},  speech recognition\cite{rezaii2019machine}, Physics\cite{porotti2019coherent,baldi2014searching}, and Geophysics \cite{de2016statistical}; the mentioned  barriers make it  impractical to exploit deep networks in portable and cheap applications.  Sparsifying a neural network can significantly reduce network parameters and learning time.  However, we need to find a suitable sparse structure which also leads to an acceptable  accuracy.

There are three main approaches to sparsifying a neural network:
\begin{enumerate}[noitemsep,topsep=0pt]
	\item Training a conventional network completely, and at the end, making it sparse with pruning techniques \cite{mengiste2015effect,han2015learning,anwar2017structured}.
	\item Modification of the loss function to enforce the sparsity \cite{scardapane2017group,louizos2017learning}.
	\item Updating a sparse structure during the learning process (evolutionary methods) \cite{mocanu2018scalable,parashar2017scnn,makhzani2013k,liu2017learning,miikkulainen2019evolving}.
\end{enumerate}

The methods using the first approach train a conventional ANN. Then, they remove the least important connections which their elimination does not decrease the accuracy of network significantly. Hence, finding a good measure to identify such connections is the most important part in this approach. Pruning in this way guarantees good performance of the   network  model \cite{han2015learning}. However, the learning time does not decrease in this approach  as  a conventional non-sparse network must be trained completely. There are also methods which fine-tune the resultant pruned  model   to compensate even small drop of accuracy in the pruning procedure\cite{sun2016sparsifying,he2017channel,han2015deep}.   
In \cite{frankle2018lottery}, it is shown that
a randomly initialized dense neural network contains a sub-network that can match the test accuracy of the
dense network and also needs the same iteration number for training (called winning
lottery ticket). However, finding this sub-network is computationally expensive as it needs multiple pruning and re-training starting
from a dense network. Thus, several references have concentrated on finding this configuration faster \cite{frankle2019stabilizing, zhou2019deconstructing, dettmers2019sparse} and using less
memory \cite{evci2020rigging}.

In the second approach, a sparsifying term is added to the loss function in order to shape the sparse structure of the network. If the loss function is chosen properly, the network tends towards
a sparse structure with good performance properties. This can be achieved by adding a sparse regularization term such as $l_1$ norm.  In MorphNet \cite{gordon2018morphnet}, this term is added to the loss function. MorphNet is a method  which iteratively
shrinks and expands  a network to find a suitable configuration. In other words, it needs multiple re-training to achieve its final configuration. 


In the third approach, both the network structure and the values of connections are trained and updated in the learning process. First of all, an initial sparse graph model is generated for the network. Then, at each epoch of the training phase, both the weights of connections and the sparse structure are updated. 
 NeuroEvolution of Augmenting Topologies (NEAT) \cite{stanley2002evolving} is an example of such methods which seeks to optimize both the weights and the topology of an ANN for a given task. Although its impressive results have been reported \cite{hausknecht2014neuroevolution,miconi2016neural}, NEAT is not  scalable to large networks as it should search a large space. 
  References \cite{verbancsics2011constraining,he2017channel,han2015deep} also propose  unscalable evolutionary methods which are not practical for large Networks.

In \cite{mocanu2018scalable}, a scalable evolutionary method is proposed, called SET. In this method, the weights are updated by back-propagation algorithm \cite{goodfellow2016deep} and the candidate connections are chosen to be replaced by new ones in each training epoch. The candidate connections are tried to be chosen from the weakest ones. Thus, replacing them leads to the  structure improvement  of the model. This method profits from the reduced learning time and acceptable performance. However, there are some drawbacks with this evolutionary method.The novelty of this method is the
process through which the structure is updated: identifying weak connections and replacing them with new ones. Hence, there are two questions to be answered: 1) what is a ``weak'' connection and
2) what makes a connection suitable for replacing the eliminated one?
 In  \cite{mocanu2018scalable}, the magnitude of
 the connections are considered as their strength (importance), which means connections with less
 magnitudes are weaker. Furthermore, new connections are chosen randomly. It seems that these
 two metrics can be chosen more efficiently by utilizing methods of evolution in natural networks. 
 
 In this paper, we proposed two evolutionary sparse methods: 1) “Path-Weight” method and
 2) “Sensitivity” method, both of which have more reasonable updating structures. The proposed
 methods show significant improvement in updating the structure compared to other existing methods. They show more generalization power  and have higher accuracy in challenging datasets like ImageNet \cite{deng2009imagenet}.  Furthermore, they need less training samples and training epochs which make them suitable for cases with low number of training samples.
 

This paper is organized as follows. In Section II and Section III, we propose the ``Path-Weight''  and the ``Sensitivity'' methods, respectively.   The simulation results are discussed in  Section IV,  and finally, we conclude the paper in section V.

\section{Path-Weight Method}\label{sec2}

As discussed earlier, an important part in the evolutionary methods for finding the sparse structures is to find the weak connections at each training epoch; this requires a measure to quantify the ``Importance'' of each connection. In reference \cite{mocanu2018scalable}, the weight magnitude of a connection  is considered as its importance.  However, in our approach, we measure the importance of a connection based on its effect on the final output. We should note two facts. First, the magnitudes of the inputs have the same importance as the weight magnitudes of the connections. A connection with small weight magnitude can be effective if the input of its node has a large magnitude; in this case, a small variation in its weight may change the output significantly. Another fact is that the ``Path'' which contains a connection is effective on its importance. We define
a path as a sequence of connections, one in each layer, which starts from the first layer and ends in the last layer. A path with large connection weight magnitudes generally has more impact on the final output of the network. A small variation in the value of the corresponding feature (feature which is the input of the path in the first layer) leads to significantly large variation
at the output of a ``strong path''. Hence, a connection with low weight magnitude in a strong path should not be eliminated in the learning procedure as by eliminating this connection, its underlying path is also eliminated which is not desirable.

By considering the aforementioned facts, an evolutionary method leading to a sparse structure called the Path-Weight method is proposed. In this method, we tend to choose the weakest connections in the weakest paths and replace them with the best candidates as opposed to random ones described in \cite{mocanu2018scalable}.

In the proposed method, the initial structure is based on Erdos-Renyi random graph\cite{erdHos1960evolution} which leads to a binomial distribution for degree of hidden neurons \cite{newman2001random}.  Let us denote  the connection between the $i^{th}$ neuron in the $k-1^{th}$ layer and the $j^{th}$ neuron in the $k^{th}$  layer by  $W_{i,j}^{(k)}$. In this initial distribution, the existence probability of connection $W_{i,j}^{(k)}$ is
\begin{equation}
P(W_{i,j}^{(k)}) = \frac{\epsilon (n^{(k)} + n^{(k-1)})}{n^{(k)}n^{(k-1)}},
\end{equation}
where $n^{(k)}$ is number of neurons in the $k^{th}$  layer and $\epsilon$ is a parameter which controls the sparsity of the model; the higher the $\epsilon$, the fewer the sparsity becomes.


In the rest of this section, the proposed method is described in three parts: identifying weak connections, adding new connections to the network in the substitution of the eliminated ones, and the time-varying version of the method.

\subsection{The Identification of Weak Connections}

 In each training epoch, the structure must be updated as well as the weights. Updating the structure needs a measure to identify the importance or the effectiveness of paths. Let us denote the $m^{th}$ path of the network by $p_m$ and  the importance measure of $p_m$ by $I_{p_m}$. This importance measure is defined as follows:
\begin{equation}\label{e2}
I_{p_m} = \prod_{l=1}^{L} I(W_l^{(p_m)}),
\end{equation}
where $L$ is the number of the network layers, and $W_l^{(p_m)}$ is the constituent connection of $p_m$ in the $l^{th}$ layer.
In \eqref{e2}, $I(W_l^{(p_m)})$ is called Normalized-Weight of $W_l^{(p_m)}$ and is defined as follows:
\begin{equation}
I(W_l^{(p_m)}) = \frac{|W_l^{(p_m)}|}{|F_l|},
\end{equation}
where $|F_l|$ denotes  the Euclidean norm of the feature vector in layer $l$.  This definition helps us to involve the effect of the input magnitude in our measure.   We choose the elimination candidates  from those paths which have smallest importance measures. Hence, the strong paths remain intact. 
For the selection of the weakest connections, we first determines a fraction $\lambda$ of the paths with the smallest importance measures and then, remove a fraction $\zeta$  of the connections with the smallest normalized-weights in those paths. $\lambda$ and $\zeta$ are the parameters which should be set beforehead.  Increasing these two parameters leads to more changes in the structure during each training epoch. This parameters also can be changed during the training procedure. At the end of this section,  we propose a time-varying version for these parameters which improves the convergence speed of the proposed method.

\subsection{Adding New Connections}
 After removing the less important connections, they should be replaced by the new ones. For selecting the best substituting candidates, we propose a probability based method which aims to add connections to the most important nodes, called key nodes, instead of randomly adding them. A key node is a node which has a significant impact in the model. In other words, removing it may ruin the performance and its connections are very important ones. If a node is a part of several strong paths, it can be a key node.  We define the importance measure of a node as the sum of the importance measures of all paths which pass through this node: 
\begin{equation}
I_{n_i^{(k)}} = \sum_{m=1}^{M}U(n_i^{(k)} \in p_m)\times I_{p_m},
\end{equation}
\begin{equation}\label{e7}
U(x) = 
\begin{cases}
1, & \text{if '$x$' is True}\\
0, & \text{otherwise}
\end{cases},
\end{equation}
where $I_{n_i^{(k)}}$ is the importance measure of $i^{th}$ node in $k^{th}$ layer  and $M$ is the total number of paths in the network. 
By normalizing this measure, we have a probability density function for adding new connections based on their source node as follows:
\begin{equation}\label{p}
\bar{I}_{n_i^{(k)}} = \frac{\delta \times I_{n_i^{(k)}}}{\sum_{i}\sum_{k}I_{n_i^{(k)}}}.
\end{equation}
Parameter $\delta$ manages the number of connections which are established in each epoch; higher values of $\delta$ leads to more established connections in each epoch. 
 Hence, a connection originating from an important node is more likely to be added.
This helps the model reach its final stable structure faster. According to ref.13, in natural networks,
adding new connections through the evolution mostly occurs in the “hub nodes” (nodes which
have more importance in connecting different network regions). Thus, our method for evolving
the network is aligned with the behavior of natural networks. The path-weight method is presented
in Algorithm \ref{a2}. 	
\begin{algorithm}[ht!]
	\label{a2}
	{
		\textbf{Initialization:}\\
		Set algorithm parameters $\epsilon$, $\lambda$, $\zeta$, and $\delta$\;
		Initialize the weights of connections by exploiting Erdos-Renyi distribution\;
		
		\textbf{Learning Procedure:}\;
		\For{each epoch}{
			Implement the standard back-propagation algorithm\;
			Update the weights\;
			\For{each connection}{
				Calculate the impotance measure\;
			}
			Detect a fraction $\lambda$ of weakest paths and then remove a fraction $\zeta$ of the connections with smaller importance measures in them\;
			\For{each node}{
				Sum the importance measures of all paths which go through the node to acheive its importance measure\;
			}
			Normalize the  importance measure of all nodes to achieve probability desnity functions for adding connections\;
			Add connections based on the probability desnity function of nodes which these connection originate from\;
		}
	}
	\caption{The Proposed Evolutionary Sparse Model}
\end{algorithm}

\subsection{ Time-Varying Version }
The parameters of the proposed methods ($\lambda$, $\zeta$, and $\delta$) directly affect on the properties of the model. These parameters can control the number of connections which are removed or established at each epoch. In the early epochs of the training, the network needs to
be changed a lot because the initialized structure is random.  On the other hand, when the number of training epochs increases, most of the important connections are established in the structure and the structure of the model reaches somehow maturity. Thus, less manipulation is required compared to the early epochs.

 In time-varying version of the proposed method, we  initialize these parameters  with the large values. 
Let us define the ``primary'' and the ``secondary'' criteria at the $t^{th}$ epoch, $C_{prim}^{(t)}$ and $C_{sec}^{(t)}$, as the following:
\begin{equation}\label{c1}
C_{prim}^{(t)} = \frac{\sum_{\forall i,k} I^{(t)}_{n_i^{(k)}}}{N},
\end{equation}
\begin{equation}\label{c2}
C_{sec}^{(t)} = \frac{\sum_{i,k \in \mho^{(t)}} I^{(t)}_{n_i^{(k)}}}{|\mho^{(t)}|},
\end{equation}
which are the average of all importance measures and the average of the eliminated ones, respectively. 	In  \eqref{c2}, $\mho^{(t)}$ is the set of eliminated connections at the $t^{th}$ epoch. 

Now, the parameters can be changed during the learning process
in the following manner:
\begin{equation}\label{e10}
\rho^{(t+1)} = 
\begin{cases}
K_1\rho^{(t)}, & \text{if   } C_{sec}<K_3 C_{prim}\\
K_2\rho^{(t)}, & \text{if   } C_{sec}>K_4 C_{prim}\\
\rho^{(t)}, & \text{otherwise}
\end{cases},
\end{equation}
where $\rho$ is $\lambda$ or $\zeta$ or $\delta$.
In\eqref{e10},  $K_1$,$K_2$,$K_3$, and $K_4$ is constants where  $K_2<1<K_1$ and $K_3<K_4<1$. According to the simulation results,  $K_1=2$, $K_2=0.5$, $K_3=0.1$, and $K_4=0.5$ are good choices for these constants.
\subsection*{  }
The simulation results show a great improvement especially in generalization power and data requirement for this method. Besides, because of the improvements in the structure updating and the convergence speed, the proposed method can reach better performance compared to non-sparse methods with fewer number of training samples. Despite its well performance, the complexity of this method may cause problems as the number of layers increases. The number of total paths in a non-sparse model exponentially increases with the increase in the number of layers. Although sparse structures have much fewer paths in comparison with the non-sparse ones, an increase in the number of layers leads to a high number of paths and, consequently, a high processing load in updating the structure. 
 Hence, in the next section, we propose another evolutionary method which has approximately no overhead process load and  has a slightly lower performance than the path-weight method.

\section{Sensitivity Method}
In this section, we propose another evolutionary method for applying sparsity in ANNs called the  sensitivity method. This  method  has low computational load despite of the path-weight method while their performance in accuracy is almost the same.   As we discussed before, the measure which determines the weak connections is so critical. In the previous section, we define  a weak connection as a connection which has small effect on the output. In other words,  there must be a large change in the weight of this connection or in its corresponding input to have a notable change at the output.   If we consider the layers of a neural network as  Multi Input- Multi Output (MIMO) systems,  the importance measure, considering the mentioned definition, can be expressed by the ``sensitivity'' measure which is defined as:  
\begin{equation}
S(W_{i,j}^{(k)}) = |\frac{\frac{\partial f}{\partial W_{i,j}^{(k)}}}{W_{i,j}^{(k)}}|,
\end{equation}
where 
$f$ stands for the final output of the neural network. This measure is well-known in Control and Systems research field and is used for measuring the importance of parameters in a system \cite{nise2020control}. 

Fortunately, the partial differentiation of the final output respect to all the connections are calculated during the training procedure by the back-propagation algorithm. Hence, there is no  overhead computational load for calculating the sensitivity measure in the updating precedure except for  dividing the partial differentiations by the corresponding weights.

The only difference between the Sensitivity method and the Path-Weight method is their
importance measures; all other stages of the methods such as initialization and adding new connections are the same. In the Sensitivity method, the importance measure of a node is considered
as the sum of sensitivity measures of all the connections that originate from that particular node:
\begin{equation}
S(n^{(k)}_i) = \sum_{j=1}^{n^{(k+1)}}|\frac{\frac{\partial f}{\partial W_{i,j}^{(k)}}}{W_{i,j}^{(k)}}|,
\end{equation}

In the training phase at each epoch, the number of required operations for the evolutionary part   is $O(L\mu\bar{N})$ for the sensitivity method; where $\bar{N}$ is the average number of the neurons in each layer and $\mu$ is the sparse factor of the model which is defined as ratio of number of existing connections to the number of all possible ones and lies between $0$ and $1$. In the other hand, the path-weight method requires $O((\mu \bar{N})^L)$ operations at each epoch. As expected, the complexity of the path-weight method grows exponentially when the number of layers increases. The linear complexity of sensitivity method makes it scalable in networks with a large number of layers. 
 Although the Path-Weight measure is more complex, the simulation
results show that the Sensitivity method performs only slightly lower than the path-weight method in
terms of accuracy, the number of required training samples, and the generalization power. Details
are discussed in the next section. 

\section{Simulation Results}
In this section, we have evaluated the performances of the proposed methods  in terms of accuracy, generalization power, convergence speed, complexity, and the number of required training samples. In the simulations, four different   networks  were compared for both multi-layer perceptron (MLP)  and convolutional  architectures.  These networks are the conventional non-sparse network, sparse evolutionary training (SET) network  (it is introduced in ref\cite{mocanu2018scalable}),  sparse network based on the path-weight method, and sparse network based on the sensitivity method.  The convolutional networks are based on ResNet18\cite{he2016deep} architecture where its last fully-connected layers is modified according to the four aforementioned methods.


In the first scenario, a MLP network is trained on CIFAR10 dataset. All networks have  $5$ layers and $2000$ neurons. They all use ReLU\cite{nair2010rectified} as the activation function. The
only differences are the number and the configuration of connections.  The dense network also uses dropout\cite{hinton2012improving} in each layer.

\begin{figure}
	\includegraphics[width=\linewidth]{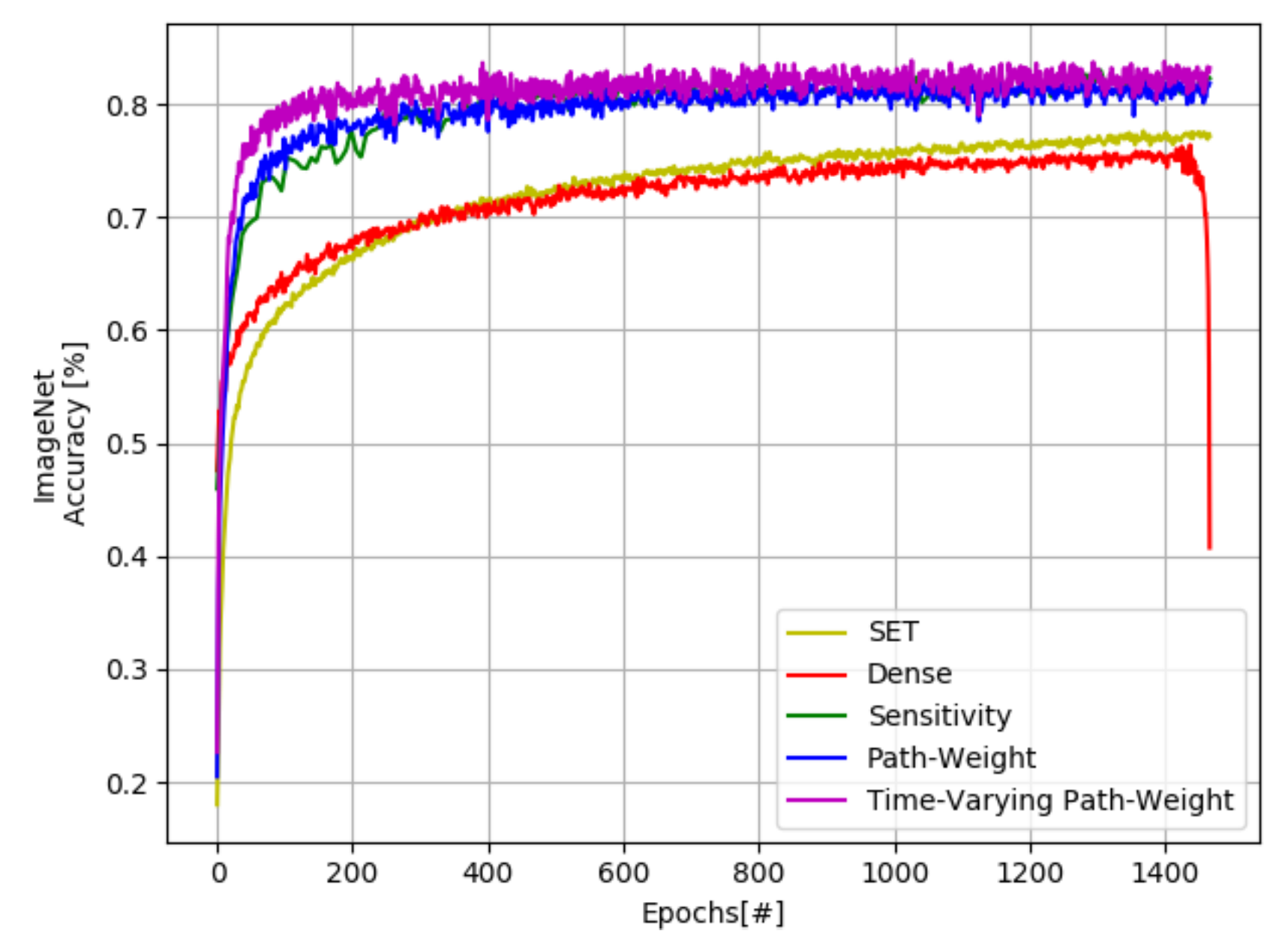}
	\caption{CFAR10 Top-1 Accuracy for Dense, SET, Path-Weight, and Sensitivity models of a MLP Network versus the epoch number.}
	\label{fig:fig1}
\end{figure}

The acuuracy of the mentioned networks versus the epoch number is shown in Fig. \ref{fig:fig1}. In this figure, the sparse networks have the tenth of possible connections which means they have $90\%$ sparsity. The networks which used proposed methods have $5\%$ better accuracy than the dense network and $3\%$ better accuracy than the SET network. The network used the path-weight method reaches $70\%$ accuracy at the $36$th epoch and the network exploited the sensitivity method reaches the same accuracy at the $60$th epoch while dense and SET networks reaches $70\%$ accuracy at $350$th epoch. This confirms that using the proposed methods can lead to an improvement in the convergence speed. The time-varying version of these methods can also accelerate the convergence speed of the model. In other words, if these parameters change with respect to the
current state of the model, a finalized structure can be achieved in fewer epochs. In Fig.1, the
time-varying version of the path-weight method reaches 80\% accuracy at 100 epochs while the other networks reach the same accuracy at epoch 500. 
Another point which can be seen from Fig. \ref{fig:fig1} is the enhancement of generalization power in the sparse networks. In other words, overfitting problem is addressed in sparse networks where accuracy of model does not drop even with thousands of training epochs. 

Convolutional Neural Networks (CNN) are the most popular networks in image related tasks. In a CNN network, last layer has MLP architecture and consequently a major fraction of parameters in the whole model. Sparsifying this layer can reduce parameters of the whole model up to 40\% which incredibly accelerates the training phase  and also improves the generalization power of the model while there is no drop in accuracy. In Fig.2, four versions of ResNet18 network for image classification of the ImageNet dataset are compared in term of accuracy at different epoch numbers.

\begin{figure}
	\includegraphics[width=\linewidth]{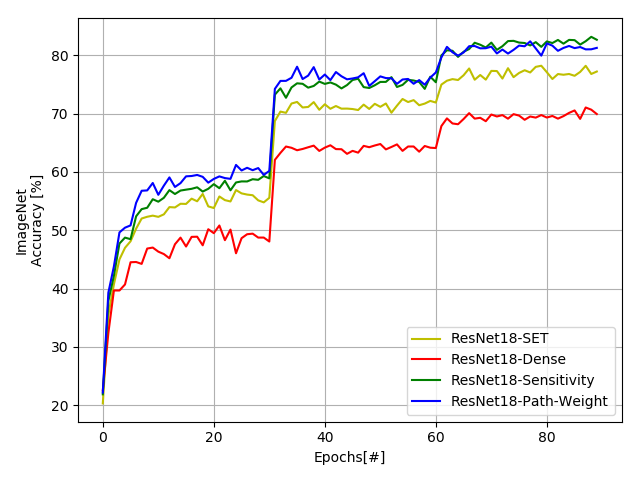}
	\caption{ImageNet top-1 accuracy for dense, SET, path-weight, and sensitivity versions of the ResNet18 network versus the epoch number.}
	\label{fig:fig2}
\end{figure}

As can be seen from Fig. \ref{fig:fig2}, the proposed methods almost have $12\%$ and $5\%$ higher top-1 accuracy in comparison with dense and SET versions of the ResNet18 network, respectively. The proposed methods have faster grow in accuracy in early epochs and their convergences to the final model happen earlier  than their sparse and non-sparse counterparts.

The proposed sparse methods  establish only important connections in the network and eliminate useless ones in order to reach a less complex model; thus, these networks can be learned with fewer number of training samples and have approximately the same performances as non-sparse ones which were trained with more samples. 

\begin{table*}[!t]
	\begin{center}
		\begin{tabular}{|c|c|c|c|c|}
			\hline
			fraction of used samples & Dense ResNet  & SET ResNet & Path-Weight ResNet & Sensitivity ResNet  \\ \hline
			100\%	&79.9\%		&77.2\%		&81.4\%		&80.5\%		\\ \hline
			90\%	&75.3\%		&74.5\%		&79.1\%		&78.3\%		\\ \hline
			80\%	&69.8\%		&68.9\%		&74.5\%		&71.4\%		\\ \hline
			60\%	&61.2\%		&65.7\%		&69.4\%		&66.1\%		\\ \hline
			40\%	&49.1\%		&57.4\%		&63.8\%		&60.3\%		\\ \hline
		\end{tabular}
	\end{center}
	\caption{The top-1 accuracy of different versions of ResNet18 for the various fraction of used training samples of the  ImageNet dataset.}
\end{table*}
\normalsize
Table 1 shows top-1 accuracy of different versions of ResNet18 for various fraction of used training samples of the  ImageNet   dataset.
This can be concluded from Table 1 that the proposed methods are more robust to the size of the training data rather than the others. This helps to use deep networks in applications such as medical imaging, cancer detection, disease prediction which there is no enough samples for conventional deep networks to be learned.

\section{Conclusion}
In this paper, we have proposed two evolutionary methods for sparsifying ANNs which update both the sparse structure
of a network and the values of its parameters during the learning procedure.  In the first method, we regard end-to-end paths of the network for updating the structure instead of regarding only single connections. In this method, which is called ``Path-Weight'' method, the effects of inputs are also considered in the updating
metric. In this approach, new connections are added to the most important nodes with a high probability.  In section 2, a less complex method is introduced
which has slightly lower properties than the ``Path-Weight'' method described in section 1.  The
sensitivity measure is used as the updating metric for this method. The simulation results show that these two methods have 5\% better accuracy, 7 times faster convergence to their final structures, and
more generalization power while they reduce the number of parameters up to 95\% in ImageNet
and CIFAR10 datasets.

\bibliographystyle{ieeetran}
\bibliography{sample}


\end{document}